\documentclass[10pt,twocolumn,letterpaper]{article}

\usepackage[pagenumbers]{wacv}      % To produce the REVIEW version for the algorithms track
\usepackage{array}
\usepackage{multirow}
\usepackage{algorithm}
\usepackage{algpseudocode}
\usepackage{soul}
\usepackage{tabularx}
\usepackage{booktabs}

\definecolor{wacvblue}{rgb}{0.21,0.49,0.74}
\usepackage[pagebackref,breaklinks,colorlinks,allcolors=wacvblue]{hyperref}

\def\wacvPaperID{1346} % *** Enter the WACV Paper ID here
\def\confName{WACV}
\def\confYear{2026}

\title{FastHMR: Accelerating Human Mesh Recovery via Token and Layer Merging with Diffusion Decoding}

\author{
\textbf{Soroush Mehraban}$^{1,2,3}$, 
\textbf{Andrea Iaboni}$^{1,3}$, 
\textbf{Babak Taati}$^{1,2,3}$\\[4pt]
$^1$University of Toronto \quad
$^2$Vector Institute \quad
$^3$KITE Research Institute, UHN\\[6pt]
\small\texttt{\href{https://soroushmehraban.github.io/FastHMR/}{Project Page: https://soroushmehraban.github.io/FastHMR/}}
}

\begin{document}
\maketitle
\begin{abstract}
Recent transformer-based models for 3D Human Mesh Recovery (HMR) have achieved strong performance but often suffer from high computational cost and complexity due to deep transformer architectures and redundant tokens. In this paper, we introduce two HMR-specific merging strategies: Error-Constrained Layer Merging (ECLM) and Mask-guided Token Merging (Mask-ToMe). ECLM selectively merges transformer layers that have minimal impact on the Mean Per Joint Position Error (MPJPE), while Mask-ToMe focuses on merging background tokens that contribute little to the final prediction. To further address the potential performance drop caused by merging, we propose a diffusion-based decoder that incorporates temporal context and leverages pose priors learned from large-scale motion capture datasets. Experiments across multiple benchmarks demonstrate that our method achieves up to 2.3$\times$ speed-up while slightly improving performance over the baseline.
\end{abstract}    
\section{Introduction}

\begin{figure}[t]
  \centering
   \includegraphics[width=1\linewidth]{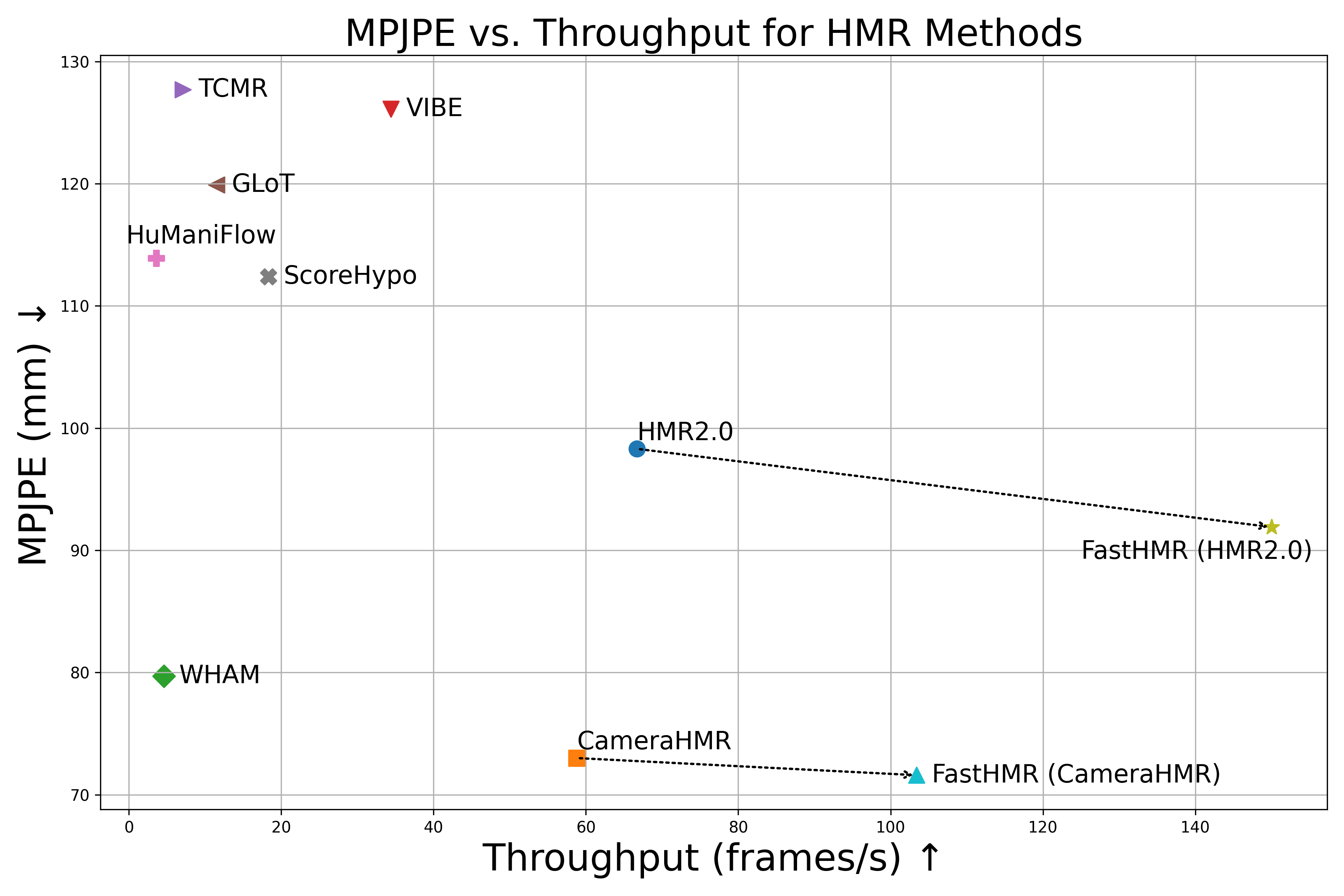}

   \caption{Throughput v.s. MPJPE error on EMDB benchmark. Throughput is evaluated on a single RTX 3090 GPU.}
   \label{fig:mpjpe-throughput}
   \vspace{-1mm}
\end{figure}

\label{sec:intro}
Human Mesh Recovery (HMR) in 3D involves estimating human pose and shape from images or videos captured by a monocular camera. Recent transformer‑based methods \cite{patel2024camerahmr, dwivedi2024tokenhmr, goel2023humans, xu2024scorehypo} have shown strong performance. However, these models require significant computational resources and memory, limiting their use in real‑time applications and on resource‑constrained hardware such as live VR/AR telepresence~\cite{habermann2019livecap}, on‑device fitness and rehabilitation, and embedded robotics. A key reason for the high computational demand is self‑attention’s quadratic complexity, which causes compute and memory to grow exponentially with sequence length and hinders real‑time or resource‑constrained HMR.

We observe that two complementary forms of redundancy persist in transformer-based HMR models: \textit{inter-layer redundancy} and \textit{spatial redundancy}. Inter-layer redundancy arises because consecutive transformer layers often learn highly correlated representations, making the full execution of all layers computationally inefficient. Spatial redundancy occurs because many tokens correspond to background regions that contribute little to the final pose or shape estimation.

Motivated by these observations, we propose two HMR-specific merging strategies: \textit{Error-Constrained Layer Merging} (ECLM) and \textit{Mask-guided Token Merging} (Mask-ToMe). ECLM identifies and merges consecutive layers whose output differences fall below an error threshold, reducing model depth while maintaining accuracy. Mask-ToMe leverages coarse person-background segmentation to merge redundant background tokens in the first transformer blocks, thereby reducing the token count while preserving essential human-centric and context information. Together, these methods significantly reduce computational overhead, but at the cost of a drop in accuracy.

To recover from the errors introduced by layer- and token-level merging, we append a diffusion-based decoder conditioned on the sequence of per-frame features produced by the merged transformer backbone. This sequence-level conditioning allows the decoder to exploit temporal context during denoising, suppressing flicker and producing smooth mesh trajectories. In addition, given that pose priors have proven effective in text-to-motion generation~\cite{chen2023executing, dai2024motionlcm}, we inject such a prior into an HMR pipeline. We achieve this by running the decoder in the latent space of a Variational Autoencoder (VAE) trained on large-scale motion-capture data, which constrains the diffusion path to anatomically plausible regions. As shown in \cref{fig:mpjpe-throughput}, this design not only restores the lost accuracy but also yields a slight improvement, while running faster than the original baseline.

In summary, our contributions are three-fold:
\begin{itemize}
    \item \textbf{Efficient architecture.} We jointly apply \emph{error-constrained layer merging} and \emph{mask-guided token merging} to accelerate inference in transformer-based HMR pipelines.
    \item \textbf{Accuracy restoration.} To offset the resulting accuracy drop, we introduce a diffusion decoder that fuses a VAE-based human-pose prior with frame-level temporal cues, yielding anatomically plausible and temporally smooth meshes.
    \item \textbf{FastHMR framework.} The resulting \textit{FastHMR} framework delivers up to 2.3$\times$ faster inference than its transformer baseline while achieving a slight improvement in estimation error.
\end{itemize}
\section{Related Works}
\label{sec:related_wrks}
\noindent\textbf{Deterministic vs.\ probabilistic estimation.}
Traditional HMR pipelines output a single mesh per frame. Optimization-based methods~\cite{smplify, pavlakos2019expressive, yang2024kitro} iteratively align SMPL~\cite{loper2023smpl} to 2-D keypoints, but depend on careful initialization. Feed-forward networks fall into two groups: image-based models~\cite{hmr, kolotouros2019learning, li2022cliff, dwivedi2024tokenhmr} and video-based variants that add temporal encoders such as 1-D CNNs \cite{kanazawa2019learning}, GRUs \cite{kocabas2020vibe}, or recurrent ViTs \cite{shin2024wham}.
To model uncertainty, recent work turns to diffusion and score-based sampling. HMDiff~\cite{foo2023distribution} starts from a random Gaussian distribution and, in addition to conditioning the denoiser using the extracted image features, it proposes a distribution alignment technique in the early diffusion steps that incorporates input-specific distribution information as a prior knowledge to simplify the denoising process. ScoreHMR~\cite{stathopoulos2024score} uses an off-the-shelf 3D human mesh regressor to estimate SMPL parameters and refines the estimation by first adding noise through DDIM Inversion~\cite{song2020denoising} and then denoising it using score guidance to align with the available observations, e.g., fitting with 2D keypoints, consistency with cross-view observations, or temporal consistency. ScoreHypo~\cite{xu2024scorehypo} proposes HypoNet network to leverage diffusion models to produce a diverse set of plausible estimates, aligned with the input image, and ScoreNet to rank them based on the quality and identify the best ones. DiffMesh~\cite{zheng2023diffmesh} is the first approach to attempt leveraging temporal information. However, it conditions diffusion on the output of MotionBERT~\cite{zhu2023motionbert}, which takes 2D pose sequences as input and lacks the complete frame context needed for generating the human mesh. DPMesh \cite{zhu2024dpmesh} is the only approach that employs an explicit generative prior, coupling diffusion with a VAE trained on \emph{RGB-image} data. In contrast, our decoder diffuses in the latent space of a VAE learned from large-scale \emph{pose} data, giving a tighter kinematic prior that allows us to recover the accuracy lost to efficiency measures while maintaining rapid inference.

\noindent\textbf{Efficient Human Mesh Recovery.} Research on efficient transformer-based Human Mesh Recovery (HMR) spans four main directions. \emph{Token pruning} techniques such as TORE discard up to $\sim$70\% of visual tokens before self-attention, cutting FLOPs while retaining accuracy~\cite{dou2023tore}. \emph{Lightweight attention} redesigns replace quadratic self-attention with cheaper variants: FastMETRO factorises METRO into a lean encoder–decoder with disentangled cross-attention for a $\times$2 throughput boost~\cite{cho2022cross}, and POTTER couples pooling attention with a high-resolution branch, shrinking parameters to 7\% of METRO without accuracy loss~\cite{zheng2023potter}. \emph{Hardware co-design} approaches such as VITA jointly tailor ViT layers and custom accelerators, reporting up-to-5$\times$ lower latency at the cost of specialized silicon~\cite{tian2024vita}. Unlike methods that rely on custom blocks, alternate modalities, or dedicated hardware, FastHMR applies error-constrained layer merging and mask-guided token merging \emph{post-hoc} to any transformer backbone and complements this compression with a diffusion decoder that restores lost accuracy while enforcing temporal smoothness. This strategy yields up to $2.3\times$ speed-ups and slight accuracy gains on standard GPUs, making FastHMR a more broadly deployable and temporally consistent solution.
\section{Method}
\subsection{Preliminaries}
\textbf{Forward Diffusion.} The forward diffusion process in latent diffusion~\cite{rombach2022high} is gradually transforms the latent vectors $Z_0$, in this context encoded pose parameters of SMPL, into noisy vectors $Z_T$ through a series of steps:
\begin{equation}
  q(Z_{1:T}|Z_0) := \prod_{t=1}^T q(Z_t|Z_{t-1}),
  \label{eq:forward-diff1}
\end{equation}
\begin{equation}
  q(Z_t|Z_{t-1}) := \mathcal{N}(Z_t;\sqrt{1-\beta_t}Z_{t-1},\beta_tI),
  \label{eq:forward-diff2}
\end{equation}
where $\{\beta_t\}_{t=1}^T$ is the variance used in the diffusion scheduler, and $T$ is number of timesteps throughout training. Using the reparametrization trick and the Markov process~\cite{ho2020denoising}, we can sample $Z_t$ in any arbitrary timestep $t$ in a closed form:
\begin{equation}
  q(Z_t|Z_0) := \sqrt{\Bar{\alpha_t}}Z_0 + \sqrt{1 - \Bar{\alpha}_t}\epsilon,
  \label{eq:forward-diff3}
\end{equation}
where $\alpha_t := 1 - \beta_t$, $\Bar{\alpha}_t := \prod_{s=1}^t\alpha_s$, and $\epsilon \sim \mathcal{N}(0, I).$ In our experiments, we use diffusion schedule with zero terminal SNR fix~\cite{lin2024common} to ensure that $\Bar{\alpha}_T = 0$.

\noindent\textbf{Backward Diffusion.} In the backward diffusion process, noisy latent vectors \( Z_T \sim \mathcal{N}(0, I) \) are sampled from a standard Gaussian distribution and gradually denoised over several steps. The denoising model \( \epsilon_\theta(Z_t, t, c) \) takes noisy latent vectors \( Z_t \) and, using the conditioning input \( c \), predicts the original noise vectors \( \epsilon \) that were added to the clean latent vectors \( Z_0 \). In our case, \( c \) contains features extracted from video frames.

\begin{algorithm}[H]
\caption{ECLM: Error-Constrained Layer Merging}\label{alg:eclm}
\begin{algorithmic}[1]
\State \textbf{Input:} pretrained HMR model $\mathcal{E}$ with $L$ layers, examples \(\mathcal{X}\), ground‐truths \(\mathcal{G}\), threshold \(\tau\)
\State \textbf{Output:} merged extractor model \(\mathcal{E}^*\)
\State \(\mathrm{MPJPE}_\mathrm{base} \gets \mathrm{ExtractMPJPE}(\mathcal{E}, \mathcal{X}, \mathcal{G})\)
\State \(high \gets L-1,\quad low \gets high - 1\)
\While{\(low \ge 0\)} 
  \State \(\mathcal{E}_{\mathrm{tmp}} \gets \mathrm{MergeLayers}(\mathcal{E}, low, high)\)
  \State \(\mathrm{MPJPE}_{\mathrm{tmp}} \gets \mathrm{ExtractMPJPE}(\mathcal{E}_{\mathrm{tmp}}, \mathcal{X}, \mathcal{G})\)
  \If{\(\mathrm{MPJPE}_{\mathrm{tmp}} - \mathrm{MPJPE}_\mathrm{base} < \tau\)}
    % \Comment{Merging low…high is acceptable}
    \State \(low \gets low - 1\)
  \Else
    \If{\(low + 1 \neq high\)}
      \State \(\mathcal{E} \gets \mathrm{MergeLayers}(\mathcal{E}, low+1, high)\)
      \State \(high \gets low\)
      \State \(low \gets high - 1\)
      % \Comment{Commit merge of layers \((low+1)\) to \(high\)}
    \Else
      \State \(high \gets high - 1\)
      \State \(low \gets low - 1\)
      % \Comment{Skip merging a single layer}
    \EndIf
  \EndIf
\EndWhile
\State \Return \(\mathcal{E}^* \gets \mathcal{E}\)
\end{algorithmic}
\end{algorithm}
\subsection{Token and Layer Merging}

We aim to reduce the computational cost in human mesh recovery (HMR). We do this by merging layers that have minimal impact on the Mean Per Joint Position Error (MPJPE), and by merging background tokens while still preserving essential person information. 

\begin{figure}[t]
  \centering
   \includegraphics[width=1\linewidth]{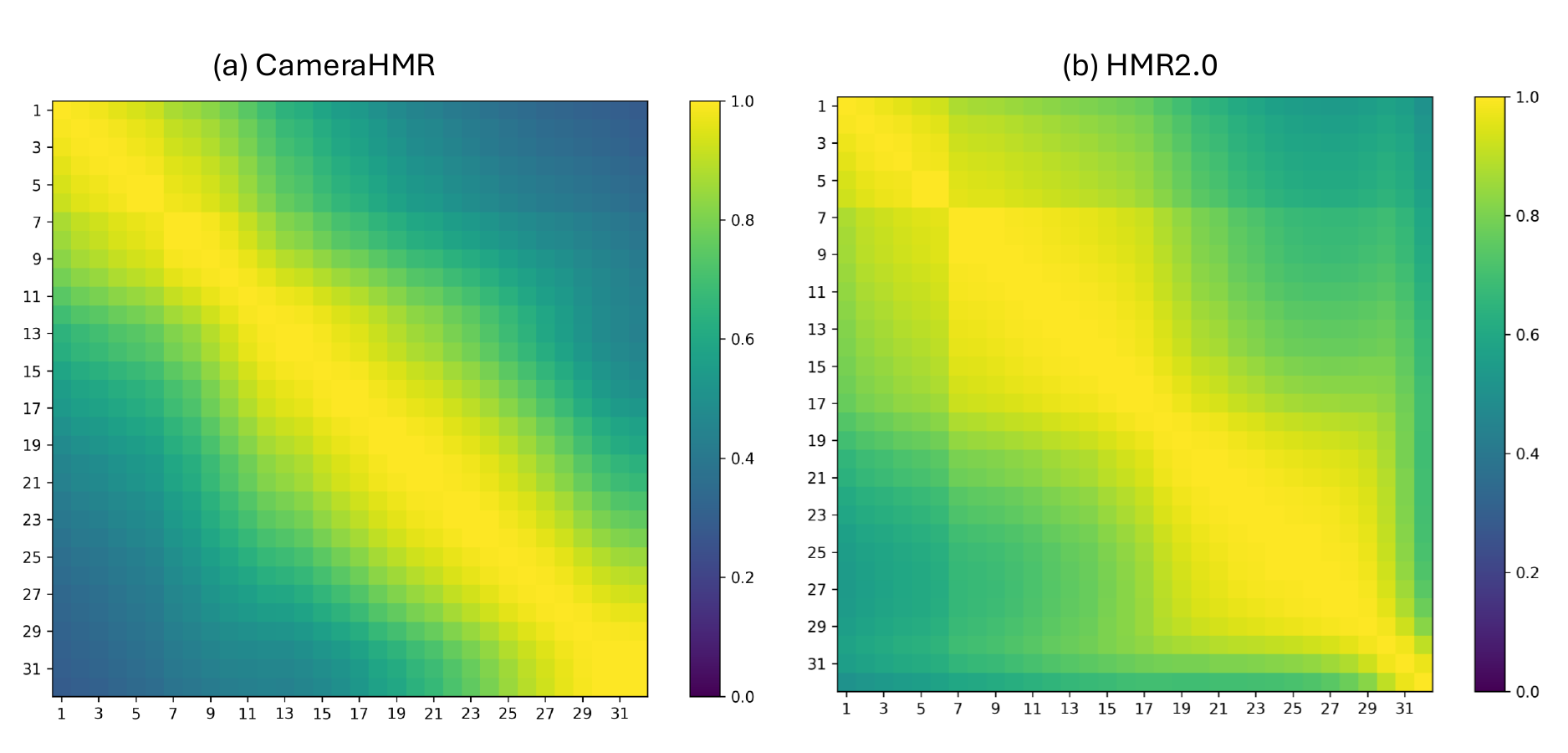}

   \caption{CKA (Center Kernel Alignment) between pairs of Transformer layers in CameraHMR~\cite{patel2024camerahmr}, and HMR2.0~\cite{goel2023humans}.}
   \label{fig:cka-hmr2}
\end{figure}

\textbf{Error-Constrained Layer Merging.} \cref{fig:cka-hmr2} shows the Centered Kernel Alignment (CKA)~\cite{ding2025sliding, raghu2021vision} scores between transformer layers in CameraHMR~\cite{patel2024camerahmr}, and HMR2.0~\cite{goel2023humans}. CKA is a metric that quantifies the similarity between internal representations of neural networks. As in large language models (LLMs)~\cite{ding2025sliding}, many layers in HMR2.0 exhibit high representational similarity. This indicates that merging such layers could reduce memory usage and inference time with minimal impact on performance. To achieve this, we propose Error-Constrained Layer Merging (ECLM). As shown in~\cref{alg:eclm}, it relies on MPJPE as a threshold to iteratively merge layers from the last layer towards the beginning. For merging, we follow SLM~\cite{ding2025sliding} and merge layers $\{L_i, L_{i+1}, ..., L_{j}\}$ with parameters $\{\theta_i, \theta_{i+1}, ..., \theta_{j}\}$ using:

\begin{equation}
\begin{split}
\theta_i^* &= \theta_i + (\theta_{i+1} - \theta_i) + \cdots + (\theta_j - \theta_i) \\
           &= \theta_i + \sum_{k=1}^{j-i} (\theta_{i+k} - \theta_i)
\end{split}
\end{equation}

\begin{figure}[t]
  \centering
   \includegraphics[width=1\linewidth]{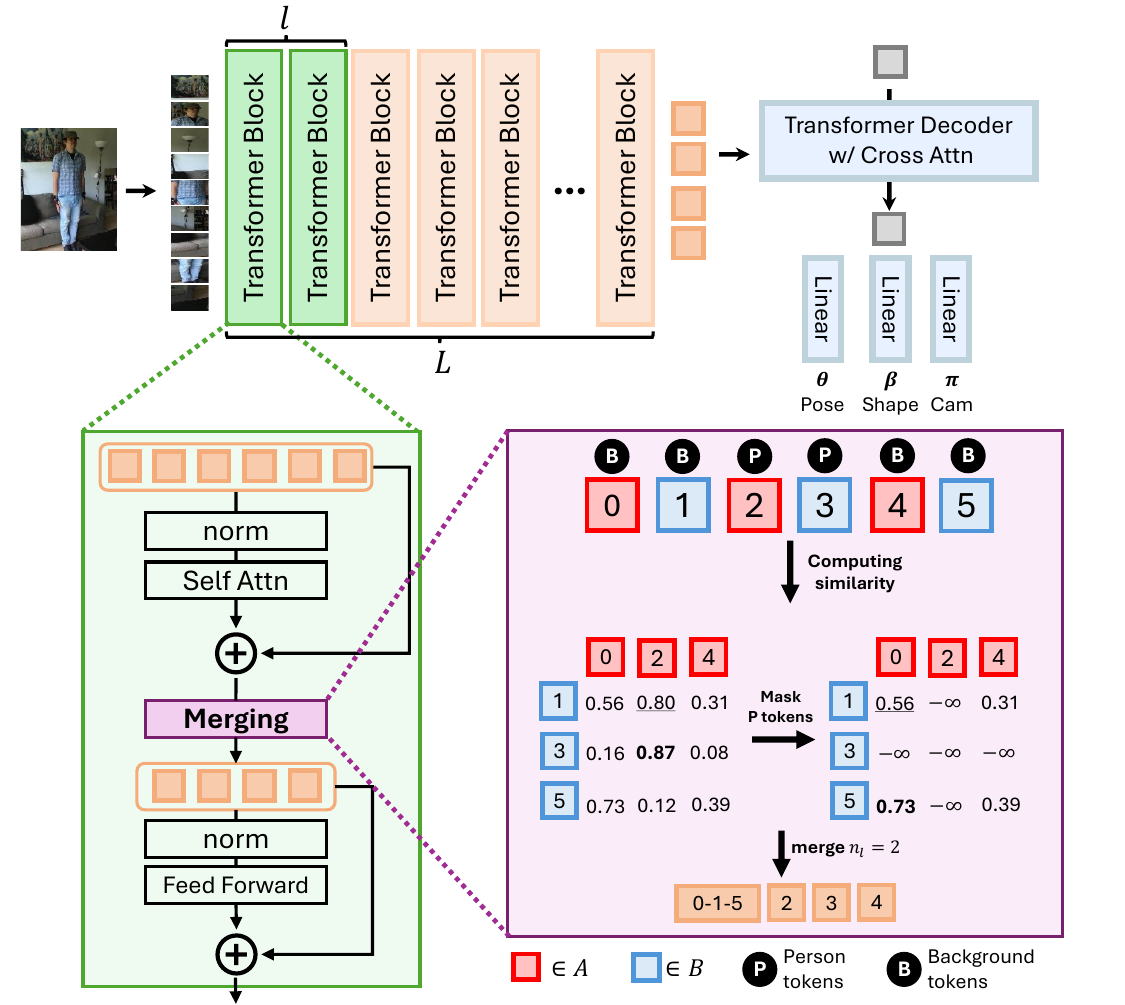}

   \caption{Overview of the Mask-ToMe strategy. Tokens are split into sets $A$ and $B$, and the most similar background token pairs are merged using similarity scores while masking out person tokens. The bold and underlined numbers represent the highest and second-highest similarity scores, respectively. The numbers shown are illustrative examples only.}
   \label{fig:mask-guided-merging}
\end{figure}
\begin{figure*}[t]
    \centering
    \includegraphics[width=\textwidth]{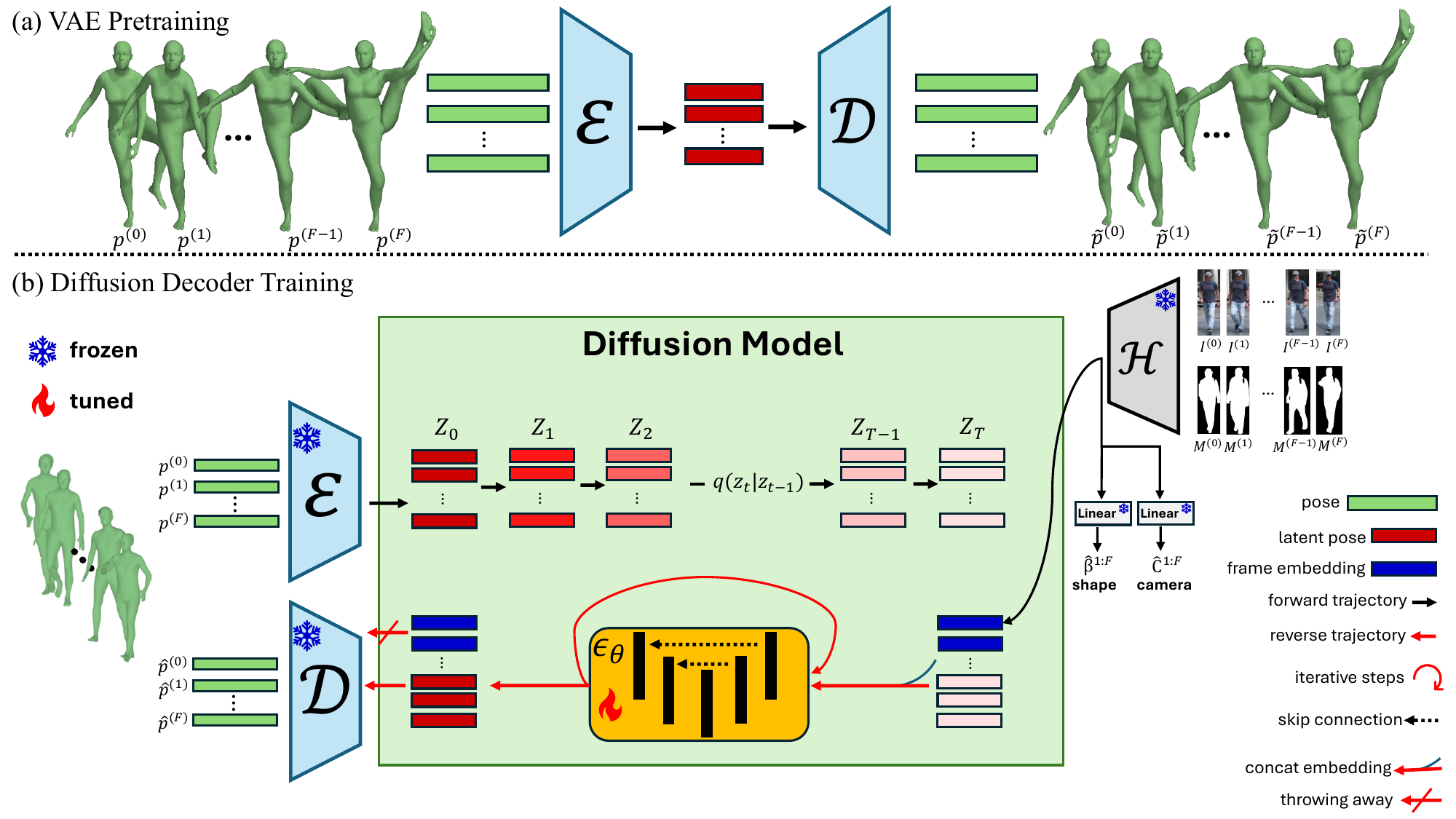}
    \caption{\textbf{Diffusion Decoder Overview.} (a) In first stage of training, a VAE model $\mathcal{V}$ is trained to learn human motion priors. (b) The second stage includes training of a denoiser $\epsilon_\theta$ to recover pose latents conditioned on per-frame encodings extracted from encoder $\mathcal{M}$.}
    \label{fig:diff-decoder}
\end{figure*}

\textbf{Mask-guided Token Merging.} Since recent HMR models~\cite{goel2023humans, patel2024camerahmr} use cross-attention in the decoder, we can safely reduce the number of tokens without affecting the final output. \cref{fig:mask-guided-merging} presents an overview of the proposed mask-guided token merging (Mask-ToMe) strategy. The input to the HMR models is a person-cropped image, which is tokenized into $N$ tokens. Among these, $M$ tokens represents the person, and the remaining $N - M$ tokens correspond to the background. Within designated transformer layers, we partition the tokens into two equal-sized sets $A$ and $B$ based on even and odd indices. Each token in set $A$ is matched to its most similar token in set $B$, from which the top $n$ most similar pairs are selected for merging by averaging. To avoid merging tokens that correspond to the person, we use a segmentation model to generate a person mask, which is tokenized alongside the image. Tokens identified as the person are assigned a similarity score of negative infinity, ensuring they are excluded from the merging process.

Among the \(L\) transformer layers, we apply token merging to the first \(l\) layers, where \(n_l\) tokens are merged at each layer. This merging follows the constraint that a fixed number of \(M'\) tokens must be retained, with \(M'\) treated as a hyperparameter. Notably, in the final merging layer \(l\), fewer than \(n_l\) tokens may be merged if necessary to satisfy this constraint. While one could use the exact number of person tokens \(M\) as a lower bound, its image-dependent variability would require sequential processing of video frames, which would significantly increase inference time.

\subsection{Diffusion Decoder}
To compensate for the performance degradation introduced by mask-guided token merging, we design a diffusion-based decoder that leverages (i) a strong pose prior learned from large-scale motion capture datasets, and (ii) the temporal dynamics of joint movements. An overview of the training pipeline of the proposed decoder is shown in~\cref{fig:diff-decoder}. It consists of a motion VAE to encode human motion prior in a lower-dimensional space and a conditional latent diffusion model to recover human mesh given the frames of an RGB video.

During inference, we use the segmentation model to track and segment a person and the resulting batch of cropped images $\mathbf{I}^{1:F} = \{I^{(i)}\}_{i=1}^F$ and corresponding segmentation masks $\mathbf{M}^{1:F} = \{M^{(i)}\}_{i=1}^F$ are given to a frame encoder $\mathcal{H}$, with our masked-guided token merging, to extract features $X^{1:F} = \{x^{(i)}\}_{i=1}^F$, where $F$ denotes the number of frames. We rely on the frame encoder $\mathcal{H}$ and its pretrained per-frame linear decoders~\cite{goel2023humans, patel2024camerahmr} to reconstruct the shape parameters $\boldsymbol{\hat{\beta}}^{1:F} = \{\hat{\beta}^{(i)}\}_{i=1}^F$ and camera parameters $\hat{C}^{1:F} = \{\hat{c}^{(i)}\}_{i=1}^F$. We do not use its linear decoder for pose and, instead, recover the pose parameters using our diffusion decoder. The latent diffusion model starts from a random Gaussian noise $\mathcal{N}(0, I)$ and, conditioned on the extracted features from frame encoder $\mathcal{H}$, outputs latent vectors $\hat{Z}_0 = \{\hat{z}_0^{(i)}\}_{i=1}^{F_Z}$, where $F_Z$ is the number of vectors in the latent space. Finally, the denoised latent vectors are passed to the VAE decoder to output pose parameters $\hat{P}^{1:F} = \mathcal{D}(Z_0)$.

\textbf{Latent Motion Representation.} As shown in~\cref{fig:diff-decoder}~(a), the motion VAE takes as input a sequence of poses $P^{1:F} = \{p^{(i)}\}_{i=1}^F$ and reconstructs it by generating a low-dimensional latent space that retains high information content. Each pose $p^{(i)} = \{J^{(i)}, \Phi^{(i)}\}$ represents the pose parameters of SMPL~\cite{loper2023smpl} by decomposing it into joint locations $J \in \mathbb{R}^{N_J \times 3}$ and twist rotations $\Phi \in \mathbb{R}^{N_\phi \times 2} = \{(cos(\phi_i), sin(\phi_i))\}_{i=1}^{N_\phi}$ based on HybrIK~\cite{li2021hybrik}, where $N_J$, $N_\phi$, and $\phi$ denote the number of body joints, number of body parts, and 1-DoF twist rotation around the $i^{th}$ body part, respectively.

The motion VAE, $\mathcal{V}=\{\mathcal{E}, \mathcal{D}\}$, employs a transformer-based architecture~\cite{petrovich2021action, chen2023executing} and consists of a transformer encoder $\mathcal{E}$ and a transformer decoder $\mathcal{D}$. It is trained to reconstruct the pose sequence $\tilde{P}^{1:F}$ by minimizing the Mean Squared Error (MSE) loss and the Kullback-Leibler (KL) loss.

\textbf{Diffusion Network.} The architecture of the proposed latent diffusion model is shown in~\cref{fig:diff-decoder}~(b). Throughout training, the ground truth (or pseudo ground truth~\cite{moon2022neuralannot}) pose $P^{1:F}$ is given to the pretrained frozen VAE encoder to output the latent pose $Z_0 = \mathcal{E}(P^{1:F})$. The forward diffusion follows~\cref{eq:forward-diff3} and constructs the noisy latent $Z_t$. For the backward process, we use the encoded per-frame features of a given video $X^{1:F} = \mathcal{H}(\mathbf{I}^{1:F})$ as the condition to guide the denoising process towards recovering the human mesh in the observed video. We keep the linear decoders for the shape and camera translation parameters but replace the pose decoder with our proposed diffusion model. Our model decodes pose parameters by analyzing the entire video as input and uses the motion prior from the first stage (VAE pretraining) to produce more accurate poses. Note that we only refine the pose parameters as previous work~\cite{kolotouros2019learning, stathopoulos2024score} shows that inferring the shape parameter $\beta$ from a single image is relatively easy and leveraging a diffusion model does not lead to any performance improvement.

The denoiser $\epsilon_\theta$ is a transformer-based denoising model with long skip connections~\cite{bao2023all, chen2023executing} that receives noisy latent vectors $Z_t \in \mathbb{R}^{N_Z \times D_Z}$ and is conditioned on per-frame features $X^{1:F} \in \mathbb{R}^{F \times D_F}$. To add the conditions as additional input, we concatenate the feature tokens $F^{1:F}$ with the noisy latent pose tokens $Z_t$. Note that the embedding size of the features ($D_F$) may differ from the size of the latent tokens ($D_Z$) and a linear projection layer is used to match them. The denoiser iteratively denoises the latent $Z_t$ until we get the estimated $\hat{Z}_0$ and decode it using $\hat{P}^{1:F} = \mathcal{D}(\hat{Z}_0)$.

To train the denoiser $\epsilon_\theta$, we rely on v-prediction~\cite{salimans2022progressive, lin2024common} objective as we empirically show that it is necessary for convergence and one-step denoising (details in ~\cref{sec:ablation-study}). More specifically, we formulate the velocity as:
\begin{equation}
  v_t = \sqrt{\Bar{\alpha}_t}\epsilon - \sqrt{1-\Bar{\alpha}_t}Z_0,
  \label{eq:v-pred}
\end{equation}
and change the denoiser to estimate velocity $\hat{v}_t = \epsilon_\theta(Z_t, t, F)$ instead of the original noise. Finally, we optimize the denoiser following a hybrid loss:
\begin{equation}
  \mathcal{L} = \lambda_1 ||\hat{v}_t - v_t||_2^2 + \lambda_2||\hat{\epsilon}_t - \epsilon_t||_2^2
  \label{eq:loss-diff},
\end{equation}
where $\hat{v}_t$ is the output of the denoiser, and $\hat{\epsilon}_t$ is derived from~\cref{eq:v-pred}. For simplicity, we set $\lambda_1 = \lambda_2 = 1$ in our experiments.
\section{Experiments}
\subsection{Datasets and Metrics}
\textbf{Datasets.} We use AMASS~\cite{mahmood2019amass} and the training split of Human3.6M~\cite{human36m} and  MPI-INF-3DHP~\cite{mehta2017monocular} to train our VAE and evaluate its reconstruction task on AMASS. For the human mesh recovery task, we train the diffusion model on Human3.6M, MPI-INF-3DHP, and BEDLAM~\cite{black2023bedlam} and evaluate it on test set of 3DPW, and EMDB (EMDB 1)~\cite{emdb} datasets. More details about the datasets are provided in the supplementary material.

\noindent\textbf{Evaluation Metrics.} To evaluate the VAE reconstruction accuracy, we compute the MPJPE using the output pose parameters and ground-truth shape parameters. For the human mesh recovery task, we compute MPJPE, Procrusted-aligned MPJPE (PA-MPJPE), and Per Vertex Error (PVE) using the pose parameter estimated by our diffusion model and shape parameter estimated from CameraHMR~\cite{patel2024camerahmr} or HMR~2.0~\cite{goel2023humans}, averaged over all the frames in a single video clip. All the errors are reported in millimeters.

\newcolumntype{?}{!{\vrule width 0.75pt}}

\begin{table*}[!t]
\centering
\setlength{\tabcolsep}{3pt}
\renewcommand{\arraystretch}{1.2}
\resizebox{0.75\textwidth}{!}
{\small{
\begin{tabular}{cl?c?cccc?cccc}
\cmidrule[0.75pt]{1-11}
& & Throughput & \multicolumn{4}{c}{3DPW (14)} & \multicolumn{4}{c}{EMDB (24)} \\

\cmidrule(lr){3-3} \cmidrule(lr){4-7} \cmidrule(lr){8-11}

& Models & \scriptsize{(frame/s)} & \scriptsize{PA-MPJPE} & \scriptsize{MPJPE} & \scriptsize{PVE} & \scriptsize{Accel} & \scriptsize{PA-MPJPE} & \scriptsize{MPJPE} & \scriptsize{PVE} & \scriptsize{Accel} \\
\cmidrule{1-11}

\multirow{9}{1em}{\rotatebox[origin=c]{90}{Deterministic}} 

& TCMR$^*$~\cite{choi2021beyond} & 7.1 & 52.7 & 86.5 & 101.4 & \underline{6.0} & 79.8 & 127.7 & 150.2 & \underline{5.3} \\
& MPS-Net$^*$~\cite{wei2022capturing}  & 22.6 & 52.1 & 84.3 & 99.0 & 6.5 & 81.4 & 123.3 & 143.9 & 6.2 \\
& VIBE$^*$~\cite{kocabas2020vibe} & 34.4 & 51.9 & 82.9 & 98.4 & 18.5 & 81.6 & 126.1 & 149.9 & 26.5 \\
& GLoT$^*$~\cite{shen2023global} & 11.5 & 50.6 & 80.7 & 96.4 & \underline{6.0} & 79.1 & 119.9 & 140.8 & 5.4 \\
& POTTER~\cite{zheng2023potter} & 83.3 & 44.8 & 75.0 & 87.4 & -- & -- & -- & -- & -- \\
& HMR~2.0$^\dagger$~\cite{goel2023humans} & 65.2 & 44.4 & 69.8 & 82.2 & 18.1 & 62.1 & 100.7 & 122.8 & 20.7 \\
& CameraHMR$^\dagger$~\cite{patel2024camerahmr} & 54.5 & 38.5 & \underline{62.1} & \underline{72.9} & 16.8 & \textbf{44.4} & \underline{73.3} & \underline{85.2} & 16.4 \\
& WHAM$^*$~\cite{shin2024wham} & 4.6 & \textbf{35.9} & \textbf{57.8} & \textbf{68.7} & 6.6 & 50.4 & 79.7 & 94.4 & \underline{5.3} \\
\cmidrule{1-11}

\multirow{7}{1em}{\rotatebox[origin=c]{90}{Probabilistic}} 
& ScoreHMR~\cite{stathopoulos2024score} & 23.4 & 51.1 & -- & -- & -- & 76.4 & -- & -- & -- \\
& HuManiFlow~\cite{sengupta2023humaniflow} ($M$=1) & 3.6 & 53.9 & 83.1 & 98.6 & -- & 76.4 & 113.9 & 133.0 & -- \\
& HMDiff$^*$($M$=25)~\cite{foo2023distribution} & -- & 44.5 & 72.7 & 82.4 & -- & -- & -- & -- & -- \\
& DiffMesh$^*$~\cite{zheng2023diffmesh} & 25.9 & 40.1 & 67.8 & 78.4 & 6.3 & -- & -- & -- & -- \\
& ScoreHypo$^*$($M$=1)~\cite{xu2024scorehypo} & 18.3 & 44.5 & 72.4 & 84.6 & -- & 77.9 & 112.4 & 131.5 & -- \\
& ScoreHypo$^*$($M$=100)~\cite{xu2024scorehypo} & 2.2 & \underline{37.6} & 63.0 & 73.4 & -- & 58.5 & 87.4 & 99.6 & -- \\
\cmidrule{2-11}
& \textbf{FastHMR-HMR2.0 ($M$=1)} & \textbf{150.0} & 41.2 & 65.8 & 78.9 & \textbf{5.4} & 53.5 & 91.9 & 104.8 & \textbf{3.4} \\
& \textbf{FastHMR-CameraHMR ($M$=1)} & \underline{103.4} & 39.1 & 62.2 & 73.9 & \textbf{5.4} & \underline{46.7} & \textbf{71.6} & \textbf{82.4} & \textbf{3.4} \\\\

\end{tabular}
}}
\vspace{-0.075in}
\caption{Quantitative comparison of models evaluated on the 3DPW \cite{3dpw} and EMDB \cite{emdb} datasets. The numbers in parentheses denote the number of body joints used to calculate MPJPE and PA-MPJPE, $M$ is the number of hypotheses, $^*$ indicates models trained including the 3DPW training set, and $^\dagger$ denotes results are reproduced using the provided checkpoints.
\textbf{Bolded} values highlight the best-performing methods for each column, while \underline{underlined} values indicate the second-best. All the errors are in $mm$. Throughput is evaluated on a single NVIDIA RTX3090 GPU.}
\label{tab:quant_cmp}
\end{table*}

\subsection{Implementation Details}
\noindent \textbf{ECLM.} We apply ECLM to the EMDB benchmark using a subset of \(n_f = 1{,}360\) frames. With a threshold of \(\tau = 0.1\,\mathrm{mm}\), the algorithm merges 6 layers in CameraHMR and 4 layers in HMR2.0.

\noindent \textbf{Mask-ToMe.} We retain \(M' = 90\) tokens and merge \(n_l = 40\) tokens across layers. As a result, the Mask-ToMe operation is applied to the first \(l = 3\) layers, with the third layer merging the remaining 10 tokens. We use YOLO11x-seg~\cite{yolo11_ultralytics} throughout training and YOLO11n-seg throughout inference time.

\noindent \textbf{Diffusion Decoder}. Similar to previous work~\cite{zhang2022mixste, zhu2023motionbert, mehraban2024motionagformer, zhao2023poseformerv2}, we use $L=243$ frames to process videos. For videos larger than 243 frames, we split them into clips and for shorter videos we resample the frames to be 243 and then resample back to the original length after processing the frames. The latent size of VAE is $N_Z \times D_Z = 27 \times 512$, equivalent with a compression ratio of 2.33:1, and the motion encoder~\cite{goel2023humans} encodes each frame into a $D_F = 1024$ dimensional vector. We pretrain the VAE for 1600 epochs using the Adam optimizer, learning rate of 0.0001, and batch size of 32. we train the diffusion model using the AdamW optimizer for 400 epochs, batch size of 32, and learning rate of 0.0001. We also use time reversing and time warping as data augmentation with a probability of 50\% throughout the diffusion training, as we found marginal improvement by including them.

All the training experiments are performed using two NVIDIA RTX 6000 GPUs, and during inference a single RTX 3090 GPU is used to assess the throughput. More details about the hyperparameters are provided in supplementary material.

\subsection{Performance Comparison}
\cref{tab:quant_cmp} compares FastHMR with other HMR methods. Although WHAM achieves the lowest error on 3DPW, it operates at only 4.6 fps due to its reliance on several heavy submodules, including ViTPose~\cite{xu2022vitpose} for 2D keypoints, the HMR2.0~\cite{goel2023humans} transformer backbone, DPVO~\cite{teed2023deep} as the SLAM module, and two RNNs. In contrast, image-based baselines such as HMR2.0 and CameraHMR achieve much higher throughput by processing frames independently. However, this per-frame inference leads to large acceleration errors, since these models lack temporal modeling and often produce jittery predictions. Probabilistic approaches involve iterative denoising and require sampling multiple noise vectors ($M > 1$) at inference time, selecting the best result for evaluation. This best-sample selection is unrealistic at deployment and, combined with the repeated sampling, results in slow runtimes (2.2–25.9 fps) that fall well below real-time. POTTER~\cite{zheng2023potter} is an efficient HMR model, achieving 83.3 fps; however, it falls behind the other models in estimation error. 

FastHMR removes this trade-off by merging backbone layers and spatial tokens once and introducing a lightweight diffusion decoder. As a result, FastHMR-HMR2.0 achieves 150.0~fps (2.3$\times$ faster than HMR2.0), and FastHMR-CameraHMR reaches 103.4~fps (1.9$\times$ faster than CameraHMR), while reducing acceleration error by roughly threefold and slightly improving estimation errors. The larger speedup observed with HMR2.0 is due to the fact that CameraHMR also relies on a HumanFoV CNN, which FastHMR does not optimize. Unlike probabilistic models that require multiple samples and iterative denoising, FastHMR relies on \emph{velocity prediction (v-prediction)} instead of noise prediction, allowing it to converge to accurate estimates with only a single denoising step and a single sample. Overall, FastHMR is the only method that simultaneously surpasses the real-time threshold, reduces temporal jitter, and preserves deterministic accuracy without relying on impractical multi-sample evaluation strategies.

\subsection{Ablation Study}
\label{sec:ablation-study}
\begin{table}
  \centering
  \begin{tabular}{@{}ccc@{}}
    \toprule
    Threshold $\tau$ (mm) & MPJPE (mm) & \# layers\\
    \midrule
    Baseline (CameraHMR) & 73.3 & 32 \\
    1.0 & 74.7 & 27 \\
    0.5 & 73.9 & 26 \\
    0.3 & 73.7 & 28 \\
    0.2 & 73.6 & 27 \\
    0.1 & 73.1 & 26 \\
    \bottomrule
  \end{tabular}
  \caption{Ablation study on MPJPE threshold $\tau$ used in ECLM. MPJPE is evaluated on EMDB benchmark.}
  \label{tab:eclm_abl}
\end{table}

\noindent \textbf{Threshold $\tau$ in ECLM.} The EMDB benchmark contains \(N_F = 24{,}117\) frames, but processing every frame through our ECLM pipeline is computationally expenrsive. To mitigate this, we uniformly sample 80 frames from each video clip in the EMDB dataset, resulting in a reduced set of \(n_f = 1{,}360\) frames. As shown in~\cref{tab:eclm_abl}, decreasing the MPJPE threshold \(\tau\) in ECLM generally leads to lower MPJPE on the EMDB benchmark, which indicates that the selected \(n_f\) frames are sufficiently representative of the full benchmark. Notably, tightening the threshold does not necessarily reduce the number of merged layers. This is because ECLM uses the threshold in a sliding window fashion, meaning that the decision to merge layers depends on local structure and may vary across different parts of the model. Interestingly, when \(\tau = 0.1\,\text{mm}\), the MPJPE slightly improves compared to the baseline, suggesting that a stricter threshold can not only preserve but also enhance accuracy.

\begin{table}
  \centering
  \begin{tabular}{@{}lcc@{}}
    \toprule
    Latent size & Reconst. Err (mm) & Recovery Err (mm) \\
    \midrule
    9$\times$256 & 6.6 & 63.1 \\
    9$\times$512 & 4.9 & 62.7 \\
    9$\times$1024 & 4.3 & 62.3 \\
    27$\times$512 & 3.6 & 62.2 \\
    \bottomrule
  \end{tabular}
  \caption{Ablation study on the VAE latent size. Reconst. Err and Recovery Err denote VAE Reconstruction Error (VAE pretraining) and Human Mesh Recovery Error (Diffusion tuning) evaluated by MPJPE on AMASS~\cite{mahmood2019amass} and 3DPW~\cite{3dpw} datasets, respectively. Both ECLM and Mask-ToMe are applied during Human Mesh Recovery and CameraHMR is used as the baseline.} 
  \label{tab:abl_vae_latent}
\end{table}

\noindent \textbf{VAE Latent Size.} \cref{tab:abl_vae_latent} investigates the impact of VAE latent dimensionality on both reconstruction quality and downstream mesh recovery performance. As the latent size increases, the VAE reconstruction error on AMASS~\cite{mahmood2019amass} decreases significantly, from 6.6\,mm for a \(9\times256\) latent to 3.6\,mm for a \(27\times512\) latent, indicating that larger latent spaces enable the VAE to capture motion details more precisely. However, the recovery error on 3DPW~\cite{3dpw} remains relatively stable around 62\,mm, with only marginal improvements at higher capacities. This suggests that the diffusion-based decoder does not fully benefit from the added expressiveness of larger latents.

\begin{table}
  \centering
  \begin{tabular}{@{}cccc@{}}
    \toprule
    Training data & PA-MPJPE & MPJPE & MVE \\
    \midrule
    w/o BEDLAM & 42.4 & 65.5 & 77.9 \\
    w/ BEDLAM & 39.1 & 62.2 & 73.9 \\
    \bottomrule
  \end{tabular}
  \caption{Ablation study on the effect of adding large-scale synthetic dataset (BEDLAM) on overall performance. All the evaluations are on FastHMR-CameraHMR and 3DPW dataset. Errors are in mm.}
  \label{tab:abl_bedlam}
\end{table}
\noindent \textbf{Training on Synthetic Data.} Adding the large‐scale synthetic BEDLAM dataset during training markedly improves FastHMR performance. As shown in~\cref{tab:abl_bedlam}, compared with the model trained without BEDLAM.
% , incorporating this extra data lowers PA‑MPJPE from 41.8\,mm to 38.4\,mm (an 8.1\% relative reduction), MPJPE from 64.9\,mm to 61.7\,mm (4.9\% reduction), and MVE from 76.9\,mm to 73.4\,mm (4.6\% reduction).
These consistent gains across all three metrics indicate that synthetic motion diversity complements limited real footage and significantly reduces the error without any architectural changes.

\begin{table}
  \centering
  \begin{tabular}{@{}ccc@{}}
    \toprule
    Training objective & PA-MPJPE & MPJPE \\
    \midrule
    Noise prediction & 97.0 & 181.8\\
    v-prediction & 39.1 & 62.5\\
    Both & 39.1 & 62.2\\
    \bottomrule
  \end{tabular}
  \caption{Ablation study on the effect of training objective on performance. All results are evaluated on FastHMR-CameraHMR using the 3DPW benchmark. Errors are reported in mm.}
  \label{tab:abl_training_objective}
\end{table}

\noindent \textbf{Training Objective.} \cref{tab:abl_training_objective} shows that adopting \emph{v}-prediction as the training objective dramatically outperforms pure noise prediction, reducing MPJPE from 181.8\,mm to 62.5\,mm and PA‑MPJPE from 97.0\,mm to 39.1\,mm. Combining the two objectives, shown in~\cref{eq:loss-diff}, yields nearly identical PA‑MPJPE to \emph{v}-prediction alone while pushing MPJPE slightly lower to 62.2\,mm. These results confirm that \emph{v}-prediction drives the major accuracy gains, and the mixed objective retains those benefits without degradation.

\begin{figure}[t]
  \centering
   \includegraphics[width=1\linewidth]{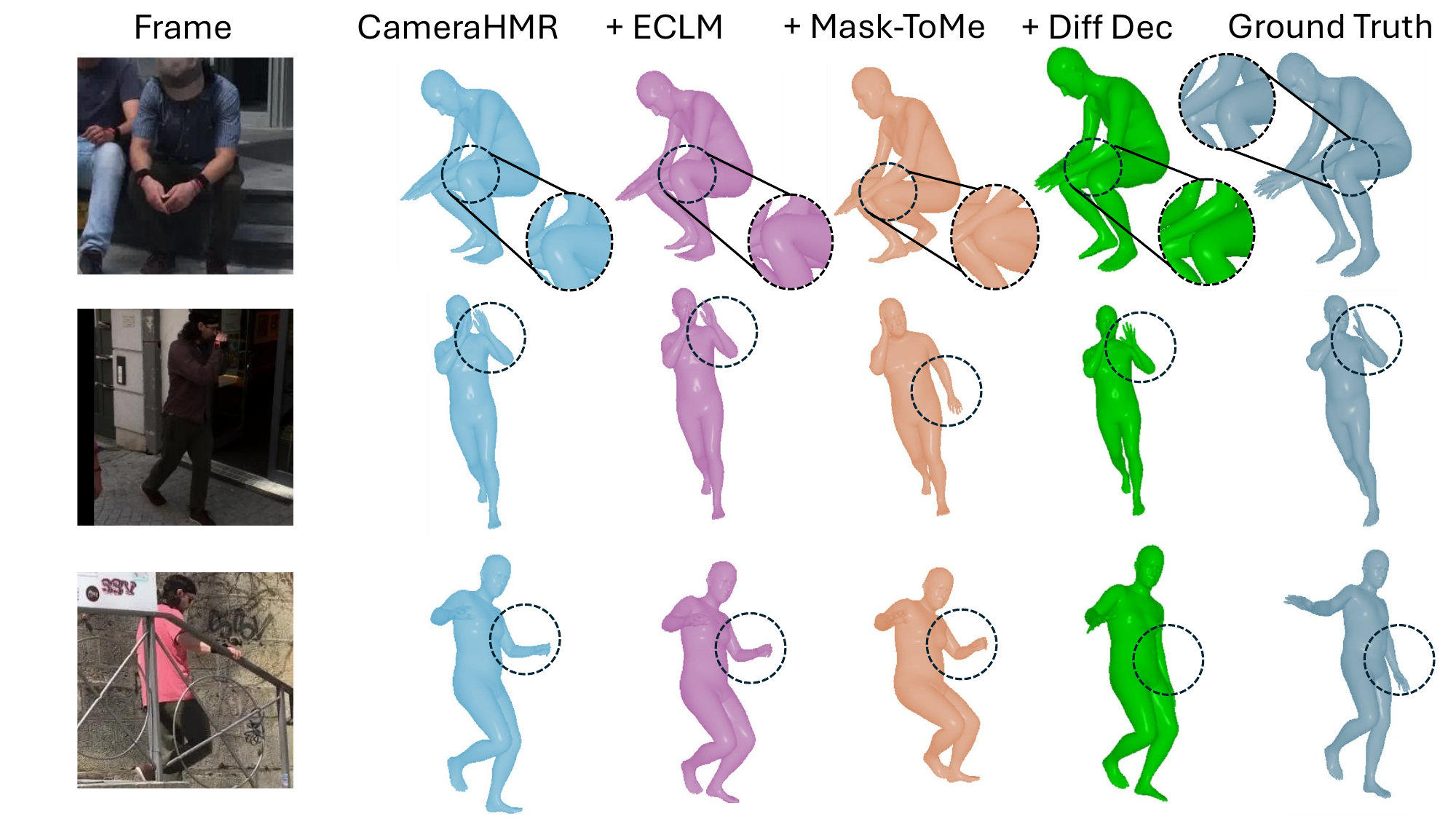}
    \caption{Qualitative comparison of mesh reconstructions across FastHMR pipeline stages.}
   \label{fig:qualitative}
\end{figure}
\noindent \textbf{Qualitative Comparison.} \cref{fig:qualitative} compares the baseline CameraHMR method with three components: ECLM, Mask-ToMe, and our diffusion decoder. ECLM has minimal impact on the estimated output mesh while improving inference speed and reducing memory usage. Mask-ToMe merges background tokens, which can sometimes contain information relevant to the human pose, and may introduce local shape errors. The diffusion decoder corrects these errors by leveraging its learned pose prior and temporal information from neighboring frames. More specifically, in the first row, Mask-ToMe distorts the lower legs, but the diffusion decoder restores a realistic shin orientation and corrects the elbow position, since elbows rarely rest between the feet in a seated pose. In the second row, excessive token merging pulls the left wrist too close to the torso; the diffusion decoder moves the hand back to its correct position by using its temporal context. In the third row, CameraHMR misplaces the self-occluded left hand behind the hip, whereas the diffusion decoder infers its forward swing from adjacent frames and aligns it much more closely with the ground truth.

\begin{figure}[t]
  \centering
   \includegraphics[width=1\linewidth]{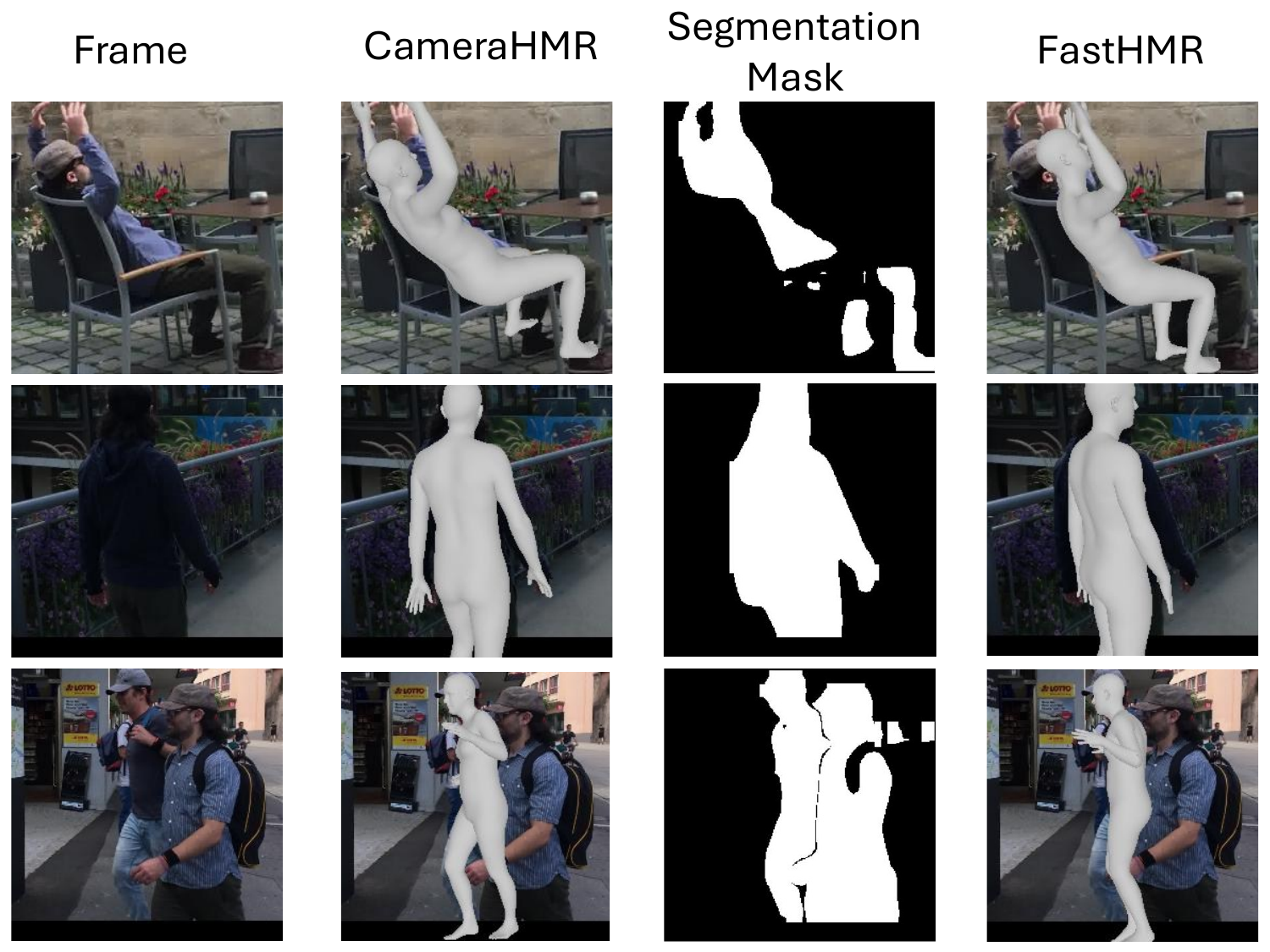}
    \caption{Failure cases of FastHMR.}
   \label{fig:failure}
\end{figure}
\noindent \textbf{Failure Cases.} \cref{fig:failure} shows failure cases of the proposed FastHMR. \emph{(i) Background tokens with semantic cues:} in the first row, although the segmentation mask correctly excludes the chair, it causes the model to classify the chair region as background. During Mask-ToMe, background tokens are aggressively merged, including the chair, which contains strong contextual cues about the human pose. As a result, the model fails to reconstruct a plausible seated pose. \emph{(ii) Low-light conditions:} in the second row, the segmentation mask is accurate, but due to poor lighting, masking out the background removes much of the usable image information. While CameraHMR benefits from global context in the background, FastHMR sees only weak body features and fails to localize the orientation. \emph{(iii) Crowded scenes with limited token capacity:} the third row includes multiple people in close proximity. Since we retain only $M'=90$ tokens, and the mask assigns many tokens to person regions, the model is forced to merge similar tokens. Some of the merged tokens belong to the subject of interest, resulting an increase in estimation error. Although this issue can be mitigated by applying a refinement step to isolate the central person throughout segmentation, we omit such heuristics for simplicity.

\begin{table}[t]
\centering
\resizebox{\linewidth}{!}{%
\begin{tabular}{@{}lccc@{}}
\toprule
\textbf{Method} & \textbf{Throughput (fps)} & \textbf{Memory (MB)} & \textbf{MPJPE (mm)} \\
\midrule
\multicolumn{4}{l}{\textbf{Baseline (CameraHMR)}} \\
\quad 2D Bbox Detection & 1500.0 & 10 & -- \\
\quad Camera Intrinsics Estimation & 333.3 & 260 & -- \\
\quad Backbone & 68.2 & 2,628 & 73.3 \\
CameraHMR (full pipeline) & 54.5 & 2,898 & 73.3 \\
\midrule
\multicolumn{4}{l}{\textbf{FastHMR}} \\
\quad Segmentation model & 1000.0 & 11 & -- \\
\quad Camera Intrinsics Estimation & 333.3 & 260 & -- \\
\quad Backbone + ECLM & 83.3 & 2,165 & 73.1 \\
\quad Backbone + ECLM + Mask-ToMe & 176.5 & -- & 84.0 \\
\quad Diffusion Decoder & 8211.0 & 351 & 71.6 \\
FastHMR-CameraHMR & 103.4 & 2,787 & 71.6 \\
\bottomrule
\end{tabular}%
}
\caption{Comparison of throughput (in frames per second), memory usage (in MB), and MPJPE (in mm) evaluated on EMDB for individual components and complete pipelines.}
\label{tab:throughput_breakdown}
\vspace{-5pt}
\end{table}

\noindent \textbf{Per-Component Analysis.} \cref{tab:throughput_breakdown} shows that FastHMR improves both speed and accuracy compared to the baseline CameraHMR. Specifically, it increases the overall throughput from 54.6\,fps to 103.4\,fps and reduces the MPJPE from 73.3\,mm to 71.6\,mm. This efficiency gain, however, does not come with reduced memory usage. The total memory consumption of FastHMR (2,787\,MB) is comparable to that of CameraHMR (2,898\,MB), indicating that while FastHMR is faster and more accurate, it does not significantly reduce memory requirements.
\section{Conclusion}\label{sec:conclusion}

We have presented FastHMR, a framework for accelerating transformer-based human mesh recovery without compromising accuracy. FastHMR employs ECLM to merge layers that have the least effect on the output and Mask-ToMe to merge spatial tokens outside the person region. To recover fine-grained detail lost during compression, a one-step diffusion decoder operating in a learned latent space refines initial predictions using its temporal and pose prior.

Extensive evaluation on 3DPW, EMDB shows that FastHMR achieves up to 2.3× speed-up on a single GPU while reducing both joint position and acceleration errors. Remaining challenges include degraded performance when segmentation masks are unreliable or background cues dominate; future work will explore adaptive merging and joint segmentation training.
{
    \small
    \bibliographystyle{ieeenat_fullname}
    \bibliography{main}
}

\clearpage
\appendix
\clearpage
\setcounter{page}{1}
\maketitlesupplementary

\appendix
\section{Datasets}
\label{sec:datasets}
\textbf{3DPW}~\cite{3dpw} is an outdoor in-the-wild dataset containing challenging scenarios, such as walking in the city and going upstairs, with ground-truth SMPL annotations. We use the test split of 3DPW for evaluation of our model.

\noindent\textbf{EMDB}~\cite{emdb} is a recently-captured dataset that recoded 10 participants in 81 indoor and outdoor environments using body-worn electromagnetic (EM) sensors and provides ground-truth SMPL parameters. It has two test splits, EMDB 1 for evaluation of 3D pose and shape in camera coordinates, and EMDB 2 for global trajectory estimation. We evaluate our model on EMDB 1 test split.

\noindent\textbf{BEDLAM}~\cite{black2023bedlam} is a large-scale synthetic dataset containing 1 million video frames with ground-truth SMPL/SMPL-X parameters. We use the train split of BEDLAM to train our network.

\noindent\textbf{Human3.6M}~\cite{human36m} is a large-scale motion capture dataset captured in indoor environment. it contains 3.6 million video frames of 11 subjects performing 15 distinct actions recorded using a single motion-capture system and 4 calibrated video cameras. Our network is trained using motion capture data from five subjects (S1, S5, S6, S7, and S8), with the data downsampled to 25 fps.

\noindent\textbf{MPI-INF-3DHP}\cite{mehta2017monocular} is a markerless dataset that spans both indoor and outdoor environments, featuring a variety of camera viewpoints, clothing styles, and human poses, along with ground-truth 3D keypoint annotations. The training dataset consists of 8 subjects, each captured in 16 videos.

\noindent\textbf{AMASS}\cite{mahmood2019amass} is a large motion-capture dataset that unifies 15 different optical marker-based mocap datasets and represent them all using SMPL~\cite{loper2023smpl} parametrization. It includes over 40 hours of motion data, covering more than 300 subjects and over 11,000 distinct motions.

\section{Additional Ablation Studies}
\noindent \textbf{Effect of swing-twist decomposition.} HybrIK~\cite{li2021hybrik} introduces an inverse kinematics approach to decompose SMPL pose parameters into joint locations and twist rotations. \cref{tab:abl_hybrik} shows that this decomposition reduces MPJPE by 33 mm, demonstrating a notable improvement in model performance. This enhancement can be attributed to two key factors. First, representing the pose parameters in the SO(3) space makes it challenging for the diffusion denoiser to denoise effectively. In SMPL’s hierarchical structure, each joint rotation is defined relative to its parent joint. Consequently, when denoising a joint rotation like the wrist, the denoiser may also adjust the parent joint, such as the elbow, to reduce error. However, this can unintentionally alter the elbow’s position from the true location. 
\begin{table}
  \centering
  \begin{tabular}{@{}cccc@{}}
    \toprule
    Data representation & PA-MPJPE & MPJPE & MVE \\
    \midrule
    w/o HybrIK & 58.1 & 95.1 & 109.2 \\
    w/ HybrIK & 37.0 & 62.1 & 72.6 \\
    \bottomrule
  \end{tabular}
  \caption{Ablation study on the effect decomposing pose parameter into joint location and twist rotation using HybrIK~\cite{li2021hybrik} method. All the evaluations are with a single-step diffusion model on 3DPW dataset. Errors are in mm.}
  \label{tab:abl_hybrik}
\end{table}
Second, since the pose parameters are defined in the SO(3) space, attributes like bone length depend on the shape parameters, which we estimate using HMR~2.0~\cite{goel2023humans} and may contain errors. Bone length, however, is critical for accurately determining joint positions. By decomposing the pose parameter into joint location and wrist rotation, the model reduces its reliance on the shape parameter, allowing it to be used for other attributes, such as body mass.

\begin{table}[t]
\centering
\begin{tabular}{@{}lcc@{}}
\toprule
\textbf{Method} & \textbf{MPJPE (3DPW)} & \textbf{MPJPE (EMDB)} \\
\midrule
w/ merging & 62.2 & 71.6 \\
w/o merging & 60.3 & 72.6 \\
\bottomrule
\end{tabular}
\caption{Impact of merging (Mask-ToMe + ECLM) on model performance after introducing the diffusion decoder. The MPJPE errors are reported in mm.}
\label{tab:diffalone}
\end{table}
\noindent \textbf{Using diffusion alone.} We propose diffusion decoder as a replacement of the naive MLP decoder used in transformer-based HMR models to enhance the robustness of model against merging methods proposed to enhance the throughput. \cref{tab:diffalone} shows that removing the merging methods from FastHMR and using only the diffusion decoder reduces MPJPE by 1.9 mm on the 3DPW dataset (used for hyperparameter tuning), but increases it by 1 mm on the EMDB benchmark (an external evaluation set). This indicates that while Mask-ToMe alone slightly worsens performance likely due to the MLP being less robust to token changes. the diffusion decoder can recover this loss. Moreover, Mask-ToMe supports the diffusion decoder by merging background tokens, which improves the model’s generalizability.

\begin{table}[t]
\centering
\begin{tabular}{@{}lcc@{}}
\toprule
\textbf{Segmentor} & \textbf{Throughput (fps)} & \textbf{MPJPE (mm)} \\
\midrule
YOLO11n-seg & 1500 & 62.2 \\
YOLO11x-seg & 300 & 61.7 \\
\bottomrule
\end{tabular}
\caption{Performance comparison when using different segmentors.}
\label{tab:segmentor_comparison}
\end{table}
\noindent \textbf{Effect of Segmentor.} \cref{tab:segmentor_comparison} shows the effect of using different segmentors in Mask-ToMe on throughput and estimation error. Although YOLO11x-seg slightly reduces the estimation error, it requires significantly more computational resources. As a result, we use YOLO11n-seg during inference in the proposed FastHMR framework.

\begin{table}[t]
\centering
\begin{tabular}{@{}lccc@{}}
\toprule
\textbf{$n_l$} & \textbf{MPJPE (3DPW)} & \textbf{MPJPE (EMDB)} \\
\midrule
Baseline & 62.1 & 73.3 \\
10 & 74.8 & 85.3 \\
20 & 74.7 & 85.0 \\
40 & 73.6 & 84.0 \\
60 & 73.1 & 83.8 \\
80 & 74.2 & 85.3 \\
100 & 80.2 & 93.5 \\
\bottomrule
\end{tabular}
\caption{The effect of merging ratio $n_l$ in Mask-ToMe, evaluated on CameraHMR model.}
\label{tab:masktome-ratio}
\end{table}
\noindent \textbf{Effect of masking ratio.} In the Mask-ToMe method, we merge $n_l$ background tokens at each layer. As shown in \cref{tab:masktome-ratio}, the choice of $n_l$ directly affects MPJPE on the 3DPW and EMDB benchmarks. In general, larger merging ratios increase the error. For example, $n_l=100$ produces the worst results. Interestingly, even smaller ratios such as 10 or 20 also lead to higher errors than expected.

\begin{figure}[t]
  \centering
   \includegraphics[width=1\linewidth]{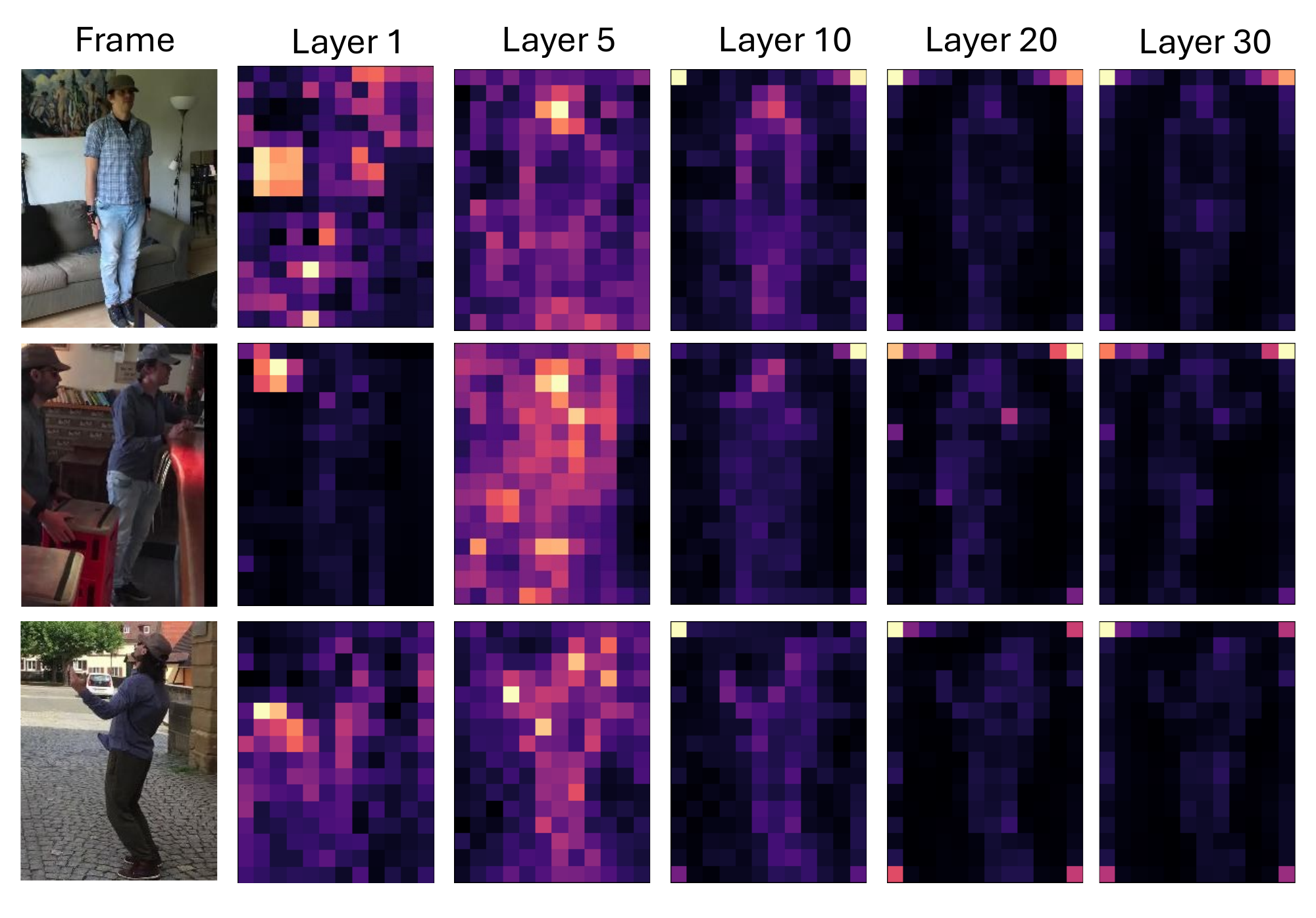}
    \caption{Attention map visualization of CameraHMR.}
   \label{fig:attn_map}
\end{figure}

To understand the underlying reason, we analyzed the attention maps (\cref{fig:attn_map}). They reveal that corner background tokens, which carry little useful information, are repurposed by the model as register tokens~\cite{darcet2023vision}. As the layers progress, the model gradually shifts important information into these tokens. When the merging ratio is too low, we risk merging them after they begin storing meaningful information, which could potentially reduce the accuracy of the final estimation.

\begin{table}
  \centering
  \begin{tabular}{@{}lccc@{}}
    \toprule
    Setting & PA-MPJPE & MPJPE & MVE \\
    \midrule
    w/o person mask & 67.2 & 98.1 & 112.9 \\
    \quad + diffusion decoder & 51.0 & 75.0 & 84.2 \\
    w/ person mask & 54.8 & 84.0 & 96.7 \\
    \quad + diffusion decoder & 46.7 & 71.6 & 82.4 \\
    \bottomrule
  \end{tabular}
  \caption{Ablation of token merging with and without person mask, and the added benefit of the diffusion decoder.}
  \label{tab:abl_mask_on_tome}
\end{table}

\noindent \textbf{Excluding person mask in Mask-ToMe} We use a person mask to merge only the background tokens while preserving the person tokens for human mesh recovery. An alternative method, proposed in ToMe~\cite{bolya2022token}, merges tokens based on similarity. \Cref{tab:abl_mask_on_tome} shows that removing the mask increases the MPJPE by 14.1~mm. Moreover, replacing the MLP decoder with a diffusion decoder does not recover this loss. These results indicate that excluding the person mask risks merging person tokens, which removes critical information for mesh recovery. Once that information is lost, even a strong decoder cannot fully compensate.

\begin{table}[t]
\centering
\resizebox{\linewidth}{!}{%
\begin{tabular}{@{}lcc@{}}
\toprule
\textbf{Method} & \textbf{Parameters (M)} & \textbf{MACs (G)} \\
\midrule
\multicolumn{3}{l}{\textbf{HMR2.0}} \\
\quad Baseline & 670.2 & 122.6 \\
\quad Baseline + ECLM & 591.5 & 107.5 \\
\quad Baseline + ECLM + Mask-ToMe & 591.5 & 52.8 \\
\midrule
\multicolumn{3}{l}{\textbf{CameraHMR}} \\
\quad Baseline & 737.1 & 144.0 \\
\quad Baseline + ECLM & 619.0 & 121.3 \\
\quad Baseline + ECLM + Mask-ToMe & 619.0 & 70.7 \\
\midrule
\quad Diffusion Decoder & 29.5 & 6.8 \\
\bottomrule
\end{tabular}%
}
\caption{Effect of proposed methods on model parameters (M) and MACs (G).}
\label{tab:param_macs}
\end{table}

\noindent \textbf{Parameters and MACs.} Table \ref{tab:param_macs} compares the impact of ECLM, Mask-ToMe, and the diffusion decoder on model complexity. For both HMR2.0 and CameraHMR, incorporating ECLM reduces parameters and multiply–accumulate operations (MACs), and combining it with Mask-ToMe yields substantial computational savings. The diffusion decoder, shown separately, is a lightweight module with relatively few parameters and MACs, highlighting its efficiency compared to the backbone models.

\begin{figure}[t]
  \centering
   \includegraphics[width=1\linewidth]{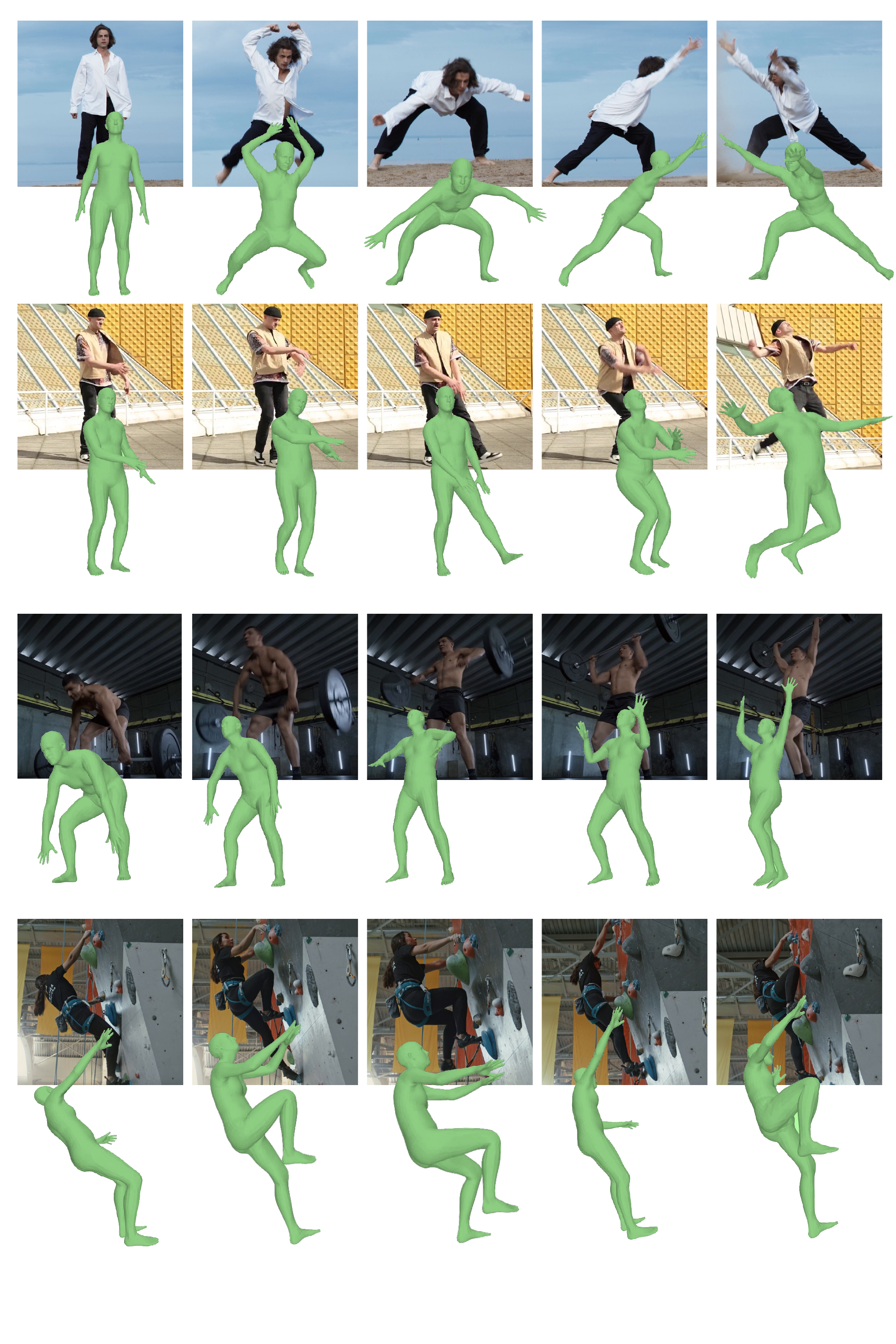}
    \caption{In-the-wild video evaluation of FastHMR}
   \label{fig:in-the-wild}
\end{figure}

\noindent \textbf{}{In-the-wild Evaluation.} ~\cref{fig:in-the-wild} shows qualitative results of FastHMR on challenging in-the-wild examples, including dynamic actions (jumping, weightlifting, climbing) and complex body articulations. Across diverse settings, our method reconstructs plausible and temporally consistent human meshes, even under fast motion, self-occlusion, and cluttered backgrounds, highlighting its robustness beyond standard benchmarks.

\section{Additional Hyperparameters}
Tables~\ref{tab:hyperparam_diff}, \ref{tab:hyperparam_denoiser}, and \ref{tab:vae_diff} denote the hyperparameters used during diffusion training, denoiser configuration, and VAE pretraining, respectively.

\section{Additional Qualitative Comparison}
\begin{figure}[t]
  \centering
   \includegraphics[width=1\linewidth]{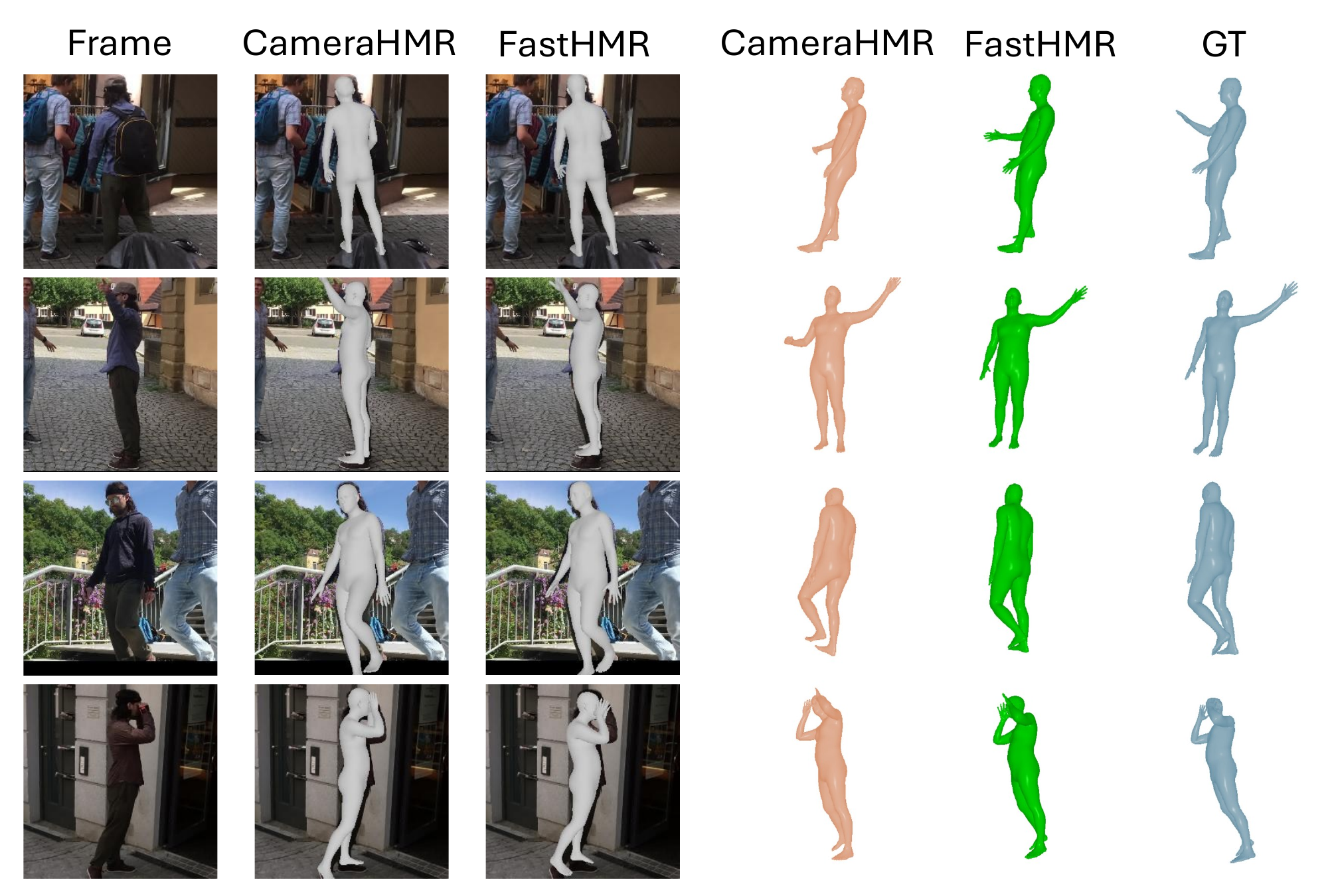}
    \caption{Qualitative comparison between CameraHMR and FastHMR-CameraHMR.}
   \label{fig:qualitative2}
\end{figure}

\begin{table}
  \centering
  \begin{tabular}{@{}cc@{}}
    \toprule
    Hyper-parameter & Value\\
    \midrule
    \# steps & 1000 \\
    $\beta_{start}$ & 0.00085 \\
    $\beta_{end}$ & 0.012 \\
    sheduler & scaled linear \\
    clip sample & False \\
    variance type & fixed small \\
    \bottomrule
  \end{tabular}
  \caption{Diffusion Hyperparameters.}
  \label{tab:hyperparam_diff}
  \vspace{-5mm}
\end{table}
\begin{table}
  \centering
  \begin{tabular}{@{}cc@{}}
    \toprule
    Hyper-parameter & Value\\
    \midrule
    Type & Transformer Encoder \\
    condition dim & 1024 \\
    embedding dim & 512 \\
    flip sin to cos & True \\
    \# frames & 243 \\
    \# encoded framed & 27 \\
    frequency shift & 0 \\
    \# heads & 4 \\
    Feedforward dim & 1024 \\
    Dropout & 0.001 \\
    Activation & GELU \\
    Normalize before & False \\
    \# Layers & 5 \\
    \bottomrule
  \end{tabular}
  \caption{Denoiser Hyperparameters.}
  \label{tab:hyperparam_denoiser}
  \vspace{-3mm}
\end{table}
\begin{table}
  \vspace{-2mm}
  \centering
  \begin{tabular}{@{}cc@{}}
    \toprule
    Hyper-parameter & Value\\
    \midrule
    input dim & $243 \times 133$ \\
    latent dim & $27 \times 512$ \\
    feedforward dim & 1024 \\
    \# layers & 9 \\
    \# heads & 4 \\
    dropout & 0.1 \\
    activation & GELU \\
    positional embedding & learned \\
    \bottomrule
  \end{tabular}
  \caption{VAE Hyperparameters.}
  \label{tab:vae_diff}
  \vspace{-5mm}
\end{table}

\cref{fig:qualitative2} compares CameraHMR and FastHMR-CameraHMR across four different frames. Although the results may appear similar depending on the camera viewpoint, CameraHMR is more likely to produce erroneous estimates for occluded body parts, whereas FastHMR demonstrates greater robustness due to its temporal awareness.

\end{document}